\pdfoutput=1

\documentclass[11pt]{article}

\usepackage{acl}

\usepackage{color, soul}
\usepackage{graphbox}
\usepackage{xcolor}
\usepackage{booktabs}

\usepackage{times}
\usepackage{latexsym}

\usepackage[T1]{fontenc}

\usepackage[utf8]{inputenc}

\usepackage{microtype}

\usepackage{url}

\usepackage{booktabs}
\usepackage{multirow}
\usepackage{subcaption}

\usepackage{amssymb} 
\usepackage{amsmath}

\usepackage{CJKutf8}

\usepackage{tikz}
\usepackage{tikz-qtree}
\usetikzlibrary{arrows,decorations.pathmorphing,backgrounds,positioning,fit,petri,shapes.misc, arrows.meta,shapes.geometric,decorations.markings,calc,shadows.blur,decorations.pathreplacing,quotes}
\definecolor{myblue}{RGB}{6, 82, 221}
\definecolor{myorange}{RGB}{211, 84, 0}
\definecolor{lowblue}{RGB}{102,178,255}
\definecolor{justblue}{RGB}{84, 160, 255}
\definecolor{mypurple}{RGB}{108, 92, 231}
\definecolor{mygray}{RGB}{158, 158, 158}
\definecolor{lowpurple}{RGB}{204,153,255}
\definecolor{lowwhite}{RGB}{255,255,255}
\definecolor{verylowpurple}{RGB}{255,102,102}
\definecolor{embcolor}{RGB}{255,255,255}
\definecolor{myred}{RGB}{235, 47, 6} 
\definecolor{mygreen}{RGB}{162, 217, 206} 
\definecolor{fontgrey}{RGB}{44, 62, 80}
\definecolor{lowpurple}{RGB}{210, 180, 222}
\definecolor{mypumpkin}{RGB}{229, 152, 102}
\definecolor{lowgreen}{RGB}{171, 235, 198}
\definecolor{lowgreen2}{RGB}{186, 220, 88}
\definecolor{lowred}{RGB}{245, 183, 177}
\definecolor{lowyellow}{RGB}{241, 196, 15}
\definecolor{mypink}{RGB}{255, 118, 117}
\definecolor{bluemartina}{RGB}{18, 203, 196}
\definecolor{puffin}{RGB}{250, 152, 58}
\definecolor{grass}{RGB}{0, 148, 50}
\definecolor{cnngray}{RGB}{116, 125, 140}

\newcommand{\squishlist}{
	\begin{list}{$\bullet$}
		{ \setlength{\itemsep}{0pt}
			\setlength{\parsep}{3pt}
			\setlength{\topsep}{3pt}
			\setlength{\partopsep}{0pt}
			\setlength{\leftmargin}{1.5em}
			\setlength{\labelwidth}{1em}
			\setlength{\labelsep}{0.5em} } }

	\newcounter{Lcount}
	\newcommand{\squishlisttwo}{
		\begin{list}{\arabic{Lcount}. }
			{ \usecounter{Lcount}
				\setlength{\itemsep}{0pt}
				\setlength{\parsep}{0pt}
				\setlength{\topsep}{0pt}
				\setlength{\partopsep}{0pt}
				\setlength{\leftmargin}{2em}
				\setlength{\labelwidth}{1.5em}
				\setlength{\labelsep}{0.5em} } }
		
		\newcommand{\squishend}{
	\end{list} }

\usepackage{multirow}
\usepackage{hhline}
\usepackage{tabularx}
\usepackage{ragged2e}
\newcolumntype{Y}{>{\RaggedRight\let\newline\\\arraybackslash\hspace{0pt}}X} 
\usepackage{xcolor}
\usepackage{pgfplots, pgfplotstable}
\pgfplotsset{compat=1.17}
\usepackage{pgfpages}
\usepackage{mathbbol}
\usepackage{arydshln}
\usepackage{graphicx}

\usepackage{amssymb}
\usepackage{amsfonts}

\usepackage{algorithm}
\usepackage{algorithmic}
\usepackage{pgf}

\usepackage{lipsum}
\usepackage{multicol}
\usepackage{changepage}

\usepackage[export]{adjustbox}

\usepackage[utf8]{inputenc}
 
\usepackage{amsthm}

\usepackage{flushend}

\title{MReD: A \underline{M}eta-\underline{Re}view \underline{D}ataset for Structure-Controllable Text Generation}

\author{
\textbf{
Chenhui Shen\thanks{$^{*}$ Equally Contributed.}~~$^\dag$\textsuperscript{\rm 1,2}~~
Liying Cheng$^{*}$\thanks{~~Chenhui, Liying, and Ran are under the Joint PhD Program between Alibaba and their corresponding universities.} \textsuperscript{\rm ~1,3}~~ 
Ran Zhou$^\dag$\textsuperscript{\rm 1,4}~~
Lidong Bing\thanks{$^\ddag$ Corresponding author.}$^\ddag$\textsuperscript{\rm 1}~~ 
Yang You\textsuperscript{\rm 2}~~
Luo Si\textsuperscript{\rm 1}}\\
\textsuperscript{\rm 1}DAMO Academy, Alibaba Group~~
\textsuperscript{\rm 2} National University of Singapore\\
\textsuperscript{\rm 3}Singapore University of Technology and Design ~~
\textsuperscript{\rm 4} Nanyang Technological University
\\
{\tt\{chenhui.shen, liying.cheng, ran.zhou\}@alibaba-inc.com} \\
{\tt\{l.bing, luo.si\}@alibaba-inc.com}
~~{\tt youy@comp.nus.edu.sg}}

\begin{document}
\maketitle

\begin{abstract}
When directly using existing text generation datasets for controllable generation, we are facing the problem of not having the domain knowledge and thus the aspects that could be controlled are limited.
A typical example is when using CNN/Daily Mail dataset for controllable text summarization, there is no guided information on the emphasis of summary sentences.
A more useful text generator should leverage both the input text and the control signal to guide the generation, which can only be built with a deep understanding of the domain knowledge.
Motivated by this vision, our paper introduces a new text generation dataset, named MReD. Our new dataset consists of 7,089 meta-reviews and all its 45k meta-review sentences are manually annotated with one of the 9 carefully defined categories, including abstract, strength, decision, etc.
We present experimental results on start-of-the-art summarization models, and propose methods for structure-controlled generation with both extractive and abstractive models using our annotated data.
By exploring various settings and analyzing the model behavior with respect to the control signal, we demonstrate the challenges of our proposed task and the values of our dataset MReD.
Meanwhile, MReD also allows us to have a better understanding of the meta-review domain.
\footnote{Our code and data are released at \url{https://github.com/Shen-Chenhui/MReD}.}

\end{abstract}

\begin{table}[t!]
	\centering
	\scalebox{0.75}{
	    \setlength{\tabcolsep}{0mm}{
\begin{tabular}{p{9.5cm}}
\midrule
    \textbf{meta-review:} \\
    \textcolor{blue}{[}This paper studies n-step returns in off-policy RL and introduces a novel algorithm which adapts the return's horizon n in function of a notion of policy's age.\textcolor{blue}{]$\leftarrow$\textsc{abstract}}
    \textcolor{blue}{[}Overall, the reviewers found that the paper presents interesting observations and promising experimental results.\textcolor{blue}{]$\leftarrow$\textsc{strength}}
    \textcolor{blue}{[}However, they also raised concerns in their initial reviews, regarding the clarity of the paper, its theoretical foundations and its positioning (notably regarding the bias/variance tradeoff of uncorrected n-step returns) and parts of the experimental results. \textcolor{blue}{]$\leftarrow$\textsc{weakness}}
    \textcolor{blue}{[}In the absence of rebuttal or revised manuscript from the authors, not much discussion was triggered.\textcolor{blue}{]$\leftarrow$\textsc{rebuttal process}}
    \textcolor{blue}{[}Based on the initial reviews, the AC cannot recommend accepting this paper, but the authors are encouraged to pursue this interesting research direction.\textcolor{blue}{]$\leftarrow$\textsc{decision}} \\
\midrule
\end{tabular}}}
    \caption{An example of annotated meta-review. \textcolor{blue}{\textsc{category}} indicates the category of each sentence.}
	\label{tab:example}
\end{table}

\section{Introduction}
Text generation entered a new era because of the development of neural network based generation techniques. Along the dimension of the mapping relation between the input information and the output text, we can roughly group the recent tasks into three clusters: more-to-less, less-to-more, and neck-to-neck. The more-to-less text generation tasks output a concise piece of text from some more abundant input, such as text summarization \cite{tan2017abstractive,kryscinski2018improving}. The less-to-more generation tasks generate a more abundant output from some obviously simpler input, such as prompt-based story generation \cite{fan2018hierarchical}. The neck-to-neck generation aims at generating an output text which conveys the same quantity of knowledge as the input but in natural language, such as typical RDF triples to text tasks \cite{gardent2017webnlg}. 

To some extent, the existing task settings are not so adequate  because they do not have a deep understanding of the domains they are working on, i.e., domain knowledge. Taking text summarization as an example, the most well-experimented dataset CNN/Daily Mail \cite{nallapati2016abstractive} is composed of the training pairs of news content and human-written summary bullets. However, it does not tell why a particular piece of news content should have that corresponding summary, for example for the same earnings report, why one media emphasizes its new business success in the summary, but another emphasizes its net income. Obviously, there is not a standard answer regarding right or wrong. For such cases, if we can specify a control signal, e.g., ``emphasizing new business'', the generated text would make more sense to users using the text generator. 

To allow controlling not only the intent of a single generated sentence but also the whole structure of a generated passage, we prepare a new dataset MReD (short for \underline{M}eta-\underline{Re}view \underline{D}ataset) with in-depth understanding of the structure of meta-reviews in a peer-reviewing system, namely the open review system of ICLR.  MReD for the first time allows a generator to be trained by simultaneously taking the text (i.e. reviews) and the structure control signal as input to generate a meta-review which is not only derivable from the reviews but also complies with the control intent. Thus from the same input text, the trained generator can generate varied outputs according to the given control signals.
For example, if the area chair is inclined to accept a borderline paper, he or she may invoke our generator with a structure of ``abstract | strength | decision'' to generate a meta-review, or may use a structure of ``abstract | weakness | suggestion'' otherwise.
Note that for ease of preparation and explanation, we ground our dataset in the peer review domain. However, the data preparation methodology and proposed models are transferable to other domains, which is indeed what we hope to motivate with this effort.

Specifically, we collect 7,089 meta-reviews of ICLR in recent years (2018 - 2021) and fully annotate the dataset.
Each sentence in a meta-review is classified into one of the 9 pre-defined intent categories: abstract, strength, weakness, rating summary, area chair (AC) disagreement, rebuttal process, suggestion, decision, and miscellaneous (misc).
Table \ref{tab:example} shows an annotated example, where each sentence is classified into a single category that best describes the intent of this sentence.
Our MReD is obviously different from the previous text generation/summarization datasets because, given the rich annotations of individual meta-review sentences, a model is allowed to learn more sophisticated generation behaviors to control the structure of the generated passage. Our proposed task is also noticeably different from the existing controllable text generation tasks (e.g., text style transfer on sentiment polarity \cite{shen2017style, liao-etal-2018-quase} and formality \cite{shang-etal-2019-semi}) because we focus on controlling the macro structure of the whole passage, rather than the wordings.


To summarize, our contributions are as follows. 
(1) We introduce a fully-annotated meta-review dataset to make better use of the domain knowledge for text generation. With thorough data analysis, we derive useful insights into the domain characteristics.
(2) We propose a new task of controllable generation focusing on controlling the passage macro structures.
It offers stronger generation flexibility and applicability for practical use cases.
(3) We design simple yet effective control methods that are independent of the model architecture.
We show the effectiveness of enforcing different generation structures with a detailed model analysis. 

\section{MReD: \underline{M}eta-\underline{Re}view \underline{D}ataset}
In this paper, we explore a new task, named the structure-controllable text generation, in a new domain, namely the meta-reviews in the peer-reviewing system.
Unlike the previous datasets that mainly focus on domains like news, the domain for meta-reviews is worth-studying because it contains essential and high-density opinions.
Specifically, during the peer review process of scientific papers, a senior reviewer or area chair will recommend a decision and manually write a meta-review to summarize the opinions from different reviews written by the reviewers. 
We first introduce the data collection process and then describe the annotation details, followed by dataset analysis.

\subsection{Data Collection}
We collect the meta-review related data of ICLR from an online peer-reviewing platform, i.e., OpenReview\footnote{\url{https://openreview.net/}} from 2018 to 2021.
Note that the submissions from earlier years are not collected because their meta-reviews are not released.
To prepare our dataset for controllable text generation, for each submission, we collect all of its corresponding official reviews with reviewer ratings and confidence scores, the final meta-review decision, and the meta-review passage.
Table \ref{tab:data1} shows the statistics of data collected from each year.
Initially, 7,894 submissions are collected.
After filtering, 7,089 meta-reviews are retained with their corresponding 23,675 reviews. Note that even without any further annotation, the dataset can already naturally serve the purpose of multi-document summarization (MDS). Compared with those conventional datasets for MDS, such as TAC \cite{owczarzak2011overview} and DUC \cite{over2004introduction}, which contain in total a few hundred input articles (equivalent to reviews in MReD), our dataset is more than 10 times larger.

\begin{table}[t!]
	\centering
    \resizebox{\columnwidth}{!}{
	    \setlength{\tabcolsep}{1.4mm}{
        \begin{tabular}{lccc}
        \toprule
        Year & \#Submissions & \#withReviews & \#Meta-Reviews \\
        \midrule
        2018 & \textcolor{white}{0,}994 & \textcolor{white}{0,}942 & \textcolor{white}{0,}892 \\
        2019 & 1,689 & 1,639 & 1,412 \\
        2020 & 2,595 & 2,517 & 2,169 \\
        2021 & 2,616 & 2,616 & 2,616 \\
        \midrule
        Total & 7,894 & 7,714 & 7,089 \\
        \bottomrule
        \end{tabular}}}
    \caption{Dataset statistics of MReD.}
	\label{tab:data1}
\end{table}

\begin{table}[t!]
	\centering
    \resizebox{\columnwidth}{!}{
	    \setlength{\tabcolsep}{1mm}{
        \begin{tabular}{p{2.8cm}p{9cm}}
        \toprule
        Categories & Definitions \\
        \midrule
        \textbf{abstract} & A piece of summary about the contents of the submission \\
        \midrule
        \textbf{strength} & Opinions about the submission's strengths \\
        \midrule
        \textbf{weakness} & Opinions about the submission's weaknesses \\
        \midrule
        \textbf{rating  summary} & A summary about reviewers' rating scores or decisions \\
        \midrule
        \textbf{ac  disagreement} & Area chair (AC) shares different opinions to reviewers \\
        \midrule
        \multirow{2}{*}{\textbf{rebuttal process}} & Contents related to authors' rebuttal with respect to reviews or discussions between reviewers in the rebuttal period \\
        \midrule
        \textbf{suggestion} & Concrete suggestions for improving the submission \\
        \midrule
        \textbf{decision} & Final decision (i.e., accept or reject) on the submission \\
        \midrule
        \textbf{miscellaneous} & None of the above, such as courtesy expressions. \\
        \bottomrule
        \end{tabular}}}
    \caption{Category definition of meta-review sentences.}
	\label{tab:category}
\end{table}

\subsection{Data Annotation}
\label{section:annotation}
As aforementioned, the structure-controllable text generation aims at controlling the structure of the generated passage. Therefore, we need to comprehensively understand the structures of meta-reviews so as to enable a model to learn how to generate outputs complying with certain structures. 

Specifically, based on the nature of meta-reviews, we pre-define 9 intent categories: abstract, strength, weakness, suggestion, rebuttal process, rating summary, area chair (AC) disagreement, decision, and miscellaneous (misc).
Table \ref{tab:category} shows the definition for each category (see example sentences in Appendix \ref{append:category_definition}).
The identification of category for some sentences is fairly straightforward, while some sentences are relatively ambiguous. Therefore, besides following the definition of each category, the annotators are also required to follow the additional rules as elaborated in Appendix \ref{append:additional_rule}

For conducting the annotation work, 14 professional data annotators from a data company are initially trained, and 12 of them are selected for the task according to their annotation quality during a trial round. 
These 12 annotators are fully paid for their work. 
Each meta-review sentence is independently labeled by 2 different annotators, and a third expert annotator resolves any disagreement between the first two annotators.
We label 45,929 sentences from 7,089 meta-reviews in total, and the Cohen’s kappa is 0.778 between the first two annotators, showing that the annotation is of quite high quality.

\subsection{Data Analysis}
\label{section:analysis}
To better understand the MReD dataset, we conduct the following analysis along different dimensions. 

\begin{figure}[t!]
\centering
\begin{tikzpicture}
\pgfplotsset{width=8cm,height=4cm,compat=1.8}
\begin{axis}[
    ybar stacked, 
    ymin=0, ymax=15000,
    symbolic x coords={abstract, strength, weakness, rating summary, rebuttal process, ac disagreement, suggestion, decision, misc},
    xtick=data,
    xticklabel style = {rotate=30, anchor=east, font=\fontsize{8}{1}\selectfont},
    yticklabel style = {font=\fontsize{8}{1}\selectfont},
    legend style={font=\fontsize{8}{1}\selectfont},
  ]
  \addplot [fill=lowred] coordinates {
  (abstract, 6000)
(strength, 2619) 
(weakness, 11004)
(rating summary, 1456) 
(rebuttal process, 3475)
(ac disagreement, 140) 
(suggestion, 2342) 
(decision, 2812)
(misc, 1983)
};
  \addplot+ [black, fill=lowgreen,
   nodes near coords, 
    nodes near coords align={anchor=south},
    every node near coord/.append style={font=\fontsize{5}{1}\selectfont}
  ] coordinates {
(abstract, 3566) 
(strength, 3058)
(weakness, 1861)
(rating summary, 859) 
(rebuttal process, 1639)
(ac disagreement, 155) 
(suggestion, 977)
(decision, 1240)
(misc, 743)
};

\legend{Reject, Accept}
\end{axis}
\end{tikzpicture}
\caption{Sentence numbers in different categories.}
\label{fig:category}
\end{figure}
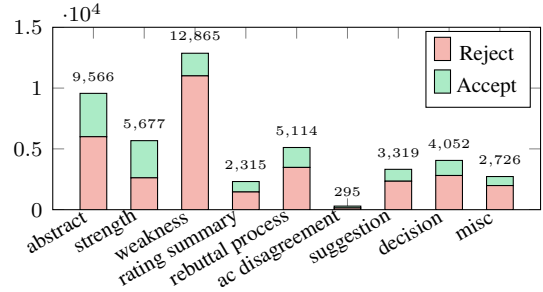

\paragraph{Sentence distribution across categories.} 
The number of sentences in different categories are shown in Figure \ref{fig:category}, breakdown by the decision (i.e., accept or reject).
Among 7,089 submissions, there are 2,368 accepted and 4,721 rejected.
Among all submissions and the rejected submissions, ``weakness'' accounts for the largest proportion,
while across the accepted ones, ``abstract'' and ``strength'' take up a great proportion.
To some extent, these three categories which dominate in meta-reviews could be easily summarized from the reviewers' comments.
However, some minor or subjective categories (e.g., ``ac disagreement'') are hard to generate.

\begin{figure}[t!]
    \center{
    \scalebox{1}{
    \includegraphics[width=\linewidth]
    {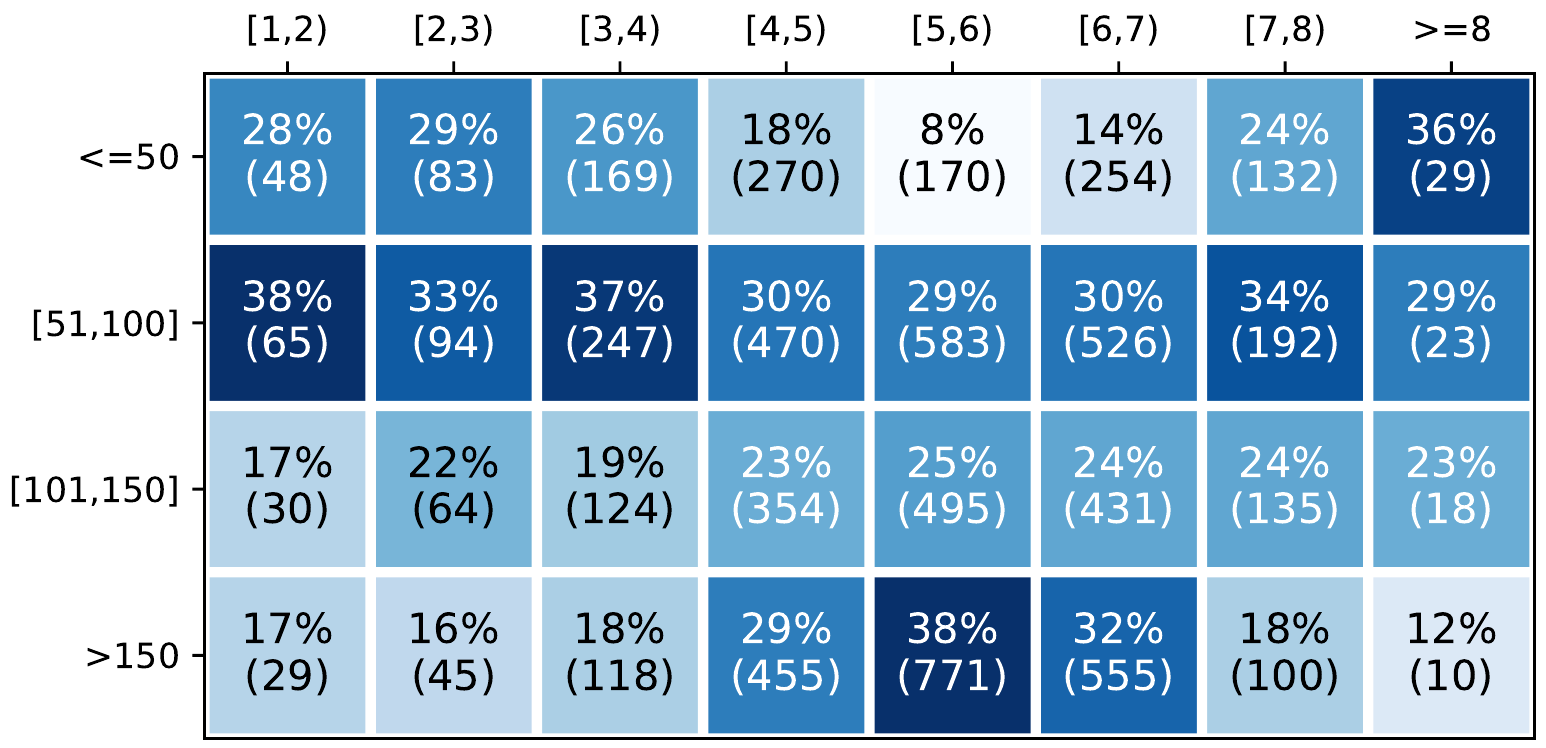}}
    }
    \caption{\label{fig:length_score}Meta-review length distribution across ratings. Bracketed numbers show the submission count.}
\end{figure}

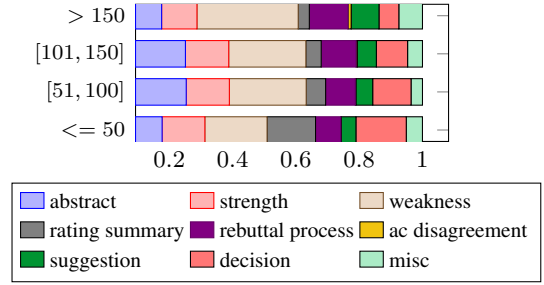
\begin{figure}[t!]
\centering{
\begin{tikzpicture}
\pgfplotsset{width=5.7cm,height=3.4cm,compat=1.8}
\begin{axis}[
    xbar stacked, 
    yticklabels={$<=50$\\ $[51, 100]$\\ $[101, 150]$\\ $>150$\\ },
    ytick=data,
    xticklabel style = {font=\fontsize{9}{1}\selectfont},
    yticklabel style = {font=\fontsize{8}{1}\selectfont},
    legend style={font=\fontsize{8}{1}\selectfont},
    legend style={at={(0.45,-0.3)},
      anchor=north,legend columns=3},
    legend cell align={left},
  ]
  \addplot  coordinates {(0.1773, 1) (0.2535, 2) (0.2510, 3) (0.1758, 4)};
  \addplot  coordinates {(0.1351, 1) (0.1363, 2) (0.1368, 3) (0.1116, 4)};
  \addplot  coordinates {(0.1965, 1) (0.2427, 2) (0.2437, 3) (0.3199, 4)};
  \addplot  coordinates {(0.1532, 1) (0.0615, 2) (0.0485, 3) (0.0352, 4)};
  \addplot  coordinates {(0.0799, 1) (0.0937, 2) (0.1102, 3) (0.1223, 4)};
  \addplot [fill=lowyellow] coordinates {(0.0007, 1) (0.0020, 2) (0.0040, 3) (0.0099, 4)};
  \addplot [fill=grass] coordinates {(0.0470, 1) (0.0530, 2) (0.0601, 3) (0.0879, 4)};
  \addplot [fill=mypink] coordinates {(0.1591, 1) (0.1212, 2) (0.0987, 3) (0.0627, 4)};
  \addplot [fill=lowgreen] coordinates {(0.0511, 1) (0.0361, 2) (0.0470, 3) (0.0747, 4)};
  
\legend{abstract, strength, weakness, rating summary, rebuttal process, ac disagreement, suggestion, decision, misc}
\end{axis}
\end{tikzpicture}}
\caption{Sentence-level category distribution percentage breakdown by different lengths of meta-reviews.}
\label{fig:length}
\end{figure}

\paragraph{Breakdown analysis by meta-review lengths and average rating scores.}
We present the percentage of meta-reviews of different lengths in each score range, as shown in Figure \ref{fig:length_score}.
For example, among the meta-reviews that receive the reviewers' average score below 2 (i.e., the first column in the figure), 28\% are less than or equal to 50 words, and 38\% fall in the length range of 51 to 100 words.
We can observe that the meta-reviews tend to be longer for those submissions receiving scores in the middle range, while shorter for those with lower scores or higher scores.
This coincides with our commonsense that for high-score and low-score submissions, the decision tends to be a clear accept or reject so that meta-reviews can be relatively shorter,
while for those borderline submissions, area chairs have to carefully weigh the pros and cons to make the final decision (see Appendix \ref{append:borderline_distri} for borderline submission analysis).
As shown in Figure \ref{fig:length}, the meta-reviews with more than 150 words generally have a larger proportion of sentences describing ``weakness'' and ``suggestion'' for authors to improve the submissions. 
Additional analysis on the category breakdown for accepted and rejected papers across the score ranges is shown in Appendix \ref{append:accept_reject_breakdown}. 

\begin{figure}[t!]
    \center{
    \includegraphics[width=1\linewidth]
    {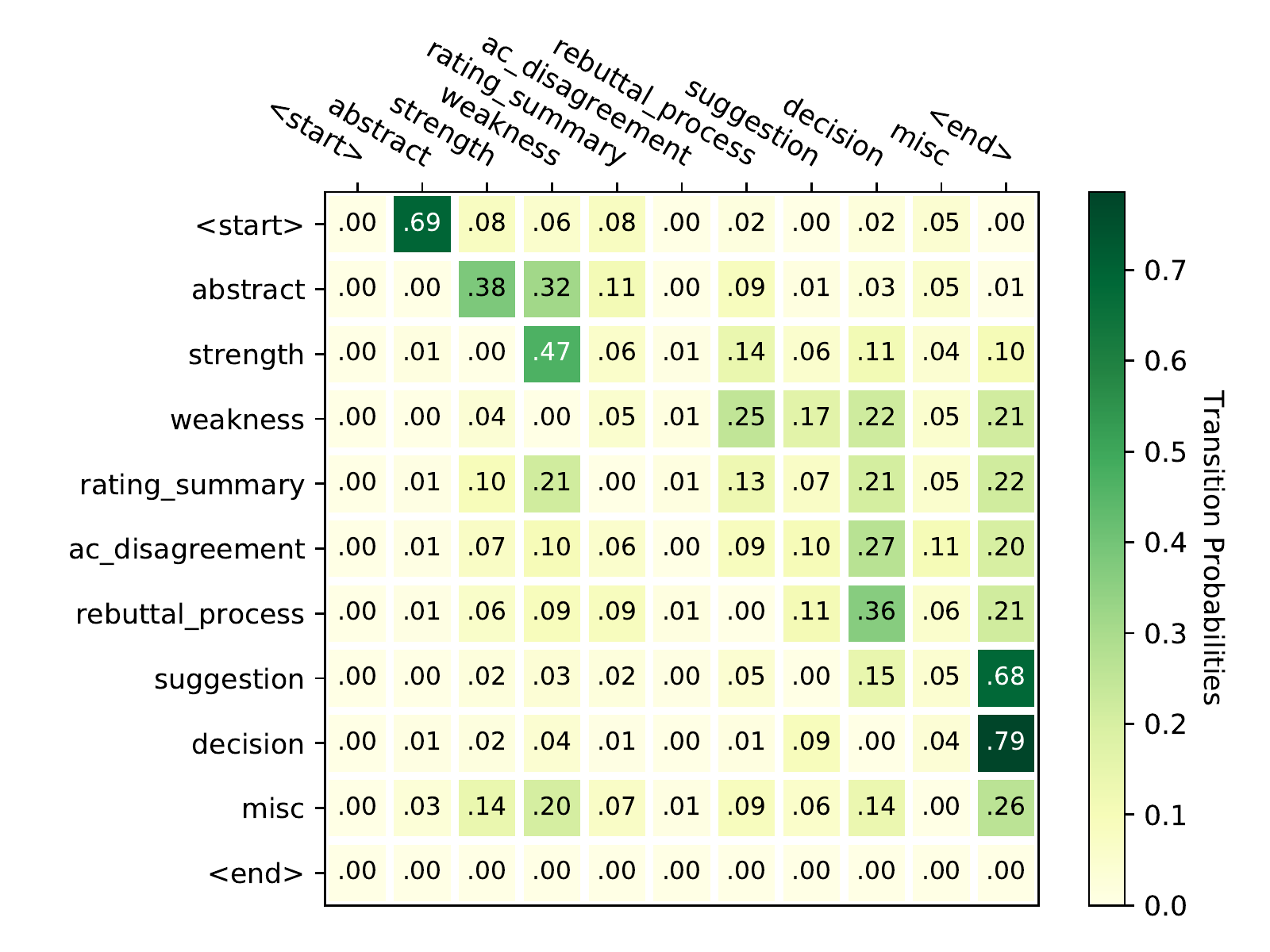}}
    \caption{Transition matrix of different categories.}
    \label{fig:transition}
\end{figure}

\paragraph{Meta-review patterns.}
To study the common structures of meta-reviews, we present the transition matrix of different category segments in Figure \ref{fig:transition}, where the sum of each row is 1.
Note that each segment represents the longest consecutive sentences with the same category.
We add ``<start>'' and ``<end>'' tokens before and after each meta-review accordingly to investigate which categories tend to be at the start/end of the meta-reviews.
It is clear to see that ``abstract'' usually positions at the beginning of the meta-review, while ``suggestion'' and ``decision'' usually appear at the end.
There are also some clear patterns appearing in the meta-reviews, such as ``abstract | strength | weakness'', ``rating summary | weakness | rebuttal process'', and ``abstract | weakness | decision''.

\section{\mbox{Structure-Controllable Text Generation}}
\label{section:gen}

\subsection{Task Definition}
As aforementioned, in uncontrolled generation, users cannot instruct the model to emphasize on desired aspects.
However, in a domain such as meta-reviews, given the same review inputs, one AC may emphasize more on the ``strength'' of the paper following a structure of ``abstract | strength | decision'', whereas another AC may prefer a different structure with more focus on reviewers' opinions and suggestions (i.e., ``rating summary'' and ``suggestion''). 
To achieve such flexibility, the task of structure-controllable text generation is defined as: given the text input (i.e., reviews) and a control sequence of the output structure, 
a model should generate a meta-review that is derivable from the reviews and presents the required structure.

\subsection{Explored Methods}
As the recent generation works~\cite{vaswani2017attention,liu2019hierarchical, xing2020automatic} basically adopt an encoder-decoder based architecture and achieve state-of-the-art performance on many tasks and datasets, we primarily investigate the performance of such a framework on our task. Thus in this subsection, we mainly present how to re-organize the input reviews and the control structure as an input sequence of the encoder. We also explore other baselines in the experiments later.

In order to summarize multiple reviews into a meta-review showing a required structure, we explicitly specify the control label sequence that a model should comply with during generation.
Specifically, we intuitively add the control sequence in front of the input text.
By directly combining both the control and textual information as a single input, our control method is independent of any specially designed encoder and decoder structures.
Moreover, by placing the short control sequence in front, an encoder can immediately observe the control signal at the very beginning, thus avoids the possible interference by the subsequent sequence. Moreover, the control sequence in front will never be truncated when the encoder truncates the input to a certain length limit.

\begin{table}[t!]
	\centering
    \resizebox{\columnwidth}{!}{
	    \setlength{\tabcolsep}{1.5mm}{
            \begin{tabular}{p{2.5cm}p{8cm}}
            \toprule
            \textbf{Combination} & \textbf{Obtained Text Input} \\
            \midrule
            \multirow{3}{*}{\textit{rate-concat}} & R1 rating score: S$_1$, R2 rating score: S$_2$, R3 rating score: S$_3$. Review1 <REVBREAK> Review2 <REVBREAK> Review3 \\
            \midrule
            \toprule
            \textbf{Control} & \textbf{Examples of Encoder Input} \\
            \midrule
            \multirow{1}{*}{\textit{sent-ctrl}}& abstract | abstract | decision ==> [TEXT INPUT] \\
            \midrule
            \textit{seg-ctrl} & abstract | decision ==> [TEXT INPUT] \\
            \midrule
            \textit{unctrl} & [TEXT INPUT] \\
            \bottomrule
            \end{tabular}
        }
    }
    \caption{Upper: example for the review combination method. 
    $S_i$ represents the score given by reviewer R$i$.
    <REVBREAK> is the special separator used to concatenate different review texts.
    Lower: examples of control methods. 
     [TEXT INPUT] refers to the obtained text from the upper section.}
	\label{tab:settings}
\end{table}


Given the multiple review inputs, we need to linearize them into a single input.
One simple method,
\textbf{\textit{concat}}, is to concatenate all inputs one after another \cite{fabbri2019multi}.
Besides the text inputs, the review rating, which cannot be found in the review passages but exists in the field of rating score, is also crucial information for writing meta-reviews.
Therefore, we create a rating sentence that consists of the extracted ratings given by the corresponding reviewers and prepend it to our concatenated review texts to obtain the final input.
We name this method \textbf{\textit{rate-concat}} (see Table \ref{tab:settings}, upper).
We also explore an alternative method, \textbf{\textit{merge}}, as follows:
From all review inputs, we use the longest one as a backbone.
We segment all reviews' content on a paragraph level, and encode them using SentenceTransformers \cite{reimers-2019-sentence-bert}.
Then, for each paragraph embedding in the non-backbone reviews, we calculate a cosine similarity score with each backbone paragraph embedding.
We then insert each non-backbone paragraph after the backbone paragraph with which it has the highest similarity score.
We repeat the process for all paragraphs in non-backbone reviews to obtain a single passage.
We further add rating sentences in front of the results of \textit{merge} to obtain \textbf{\textit{rate-merge}}.
Additionally, we provide a  \textbf{\textit{longest-review}} baseline, which does not combine reviews but only uses the longest review as the input.

As aforementioned, we place the control sequence in front of the re-organized review information. 
Specifically, we explore two different control methods, namely, \textbf{\textit{sent-ctrl}} and \textbf{\textit{seg-ctrl}}. \textit{Sent-ctrl} uses one control label per target sentence and controls generation on the sentence-level. 
Note that this method can allow implicit control on the length (i.e., number of sentences) of the generation.  
\textit{Seg-ctrl} treats consecutive sentences of the same label as one segment and only 
uses one label for a single segment.
Example inputs of 
different control settings are shown in Table \ref{tab:settings} (lower).
For instance, sent-ctrl repeats ``abstract'' in its control sequence whereas seg-ctrl does not. This is because seg-ctrl treats the 1\textsuperscript{st} and 2\textsuperscript{nd} target sentences of ``abstract'' as the same segment and only uses a single label to indicate it in the sequence.
Additionally, we provide a vanilla setting for uncontrolled generation, \textbf{\textit{unctrl}}, where no control sequence is used.

Using the above input sequence as the source and the corresponding meta-review as the target, we can train an encoder-decoder model for controllable generation.
Many transformer-based models have achieved state-of-the-art performance.
Common abstractive summarization models include BART \cite{lewis2020bart}, T5 \cite{raffel2020exploring} and PEGASUS \cite{zhang2020pegasus}.
In this paper we focus on the \textit{bart-large-cnn} model, one variant of the BART model (results on other pretrained models can be found in Appendix \ref{append:pretrained-models}, which show similar trend).
More specifically, we use the PyTorch implementation in the open-source library Hugging Face Transformers \cite{wolf2020transformers} and its hosted pretrained models\footnote{\url{https://huggingface.co/models}}.

\section{Experiments}

\subsection{Baselines}
\paragraph{Extractive Baselines.}
We employ three common extractive summarization baselines each of which basically provides a mechanism to rank the input sentences. \textbf{\textit{LexRank}} \cite{erkan2004lexrank} represents sentences in a graph and uses eigenvector centrality to calculate sentence importance scores. \textbf{\textit{TextRank}} \cite{mihalcea2004textrank} is another graph-based sentence ranking method that obtains vertex scores by running a ``random-surfer model'' until convergence.  \textbf{\textit{MMR}} \cite{carbonell1998use} calculates sentence scores by 
balancing the redundancy score with the information relevance score.
After ranking with each of the above models, we select sentences as output with different strategies according to the controlled and uncontrolled settings. 
For the uncontrolled setting, we simply select the top $k$ sentences as the generated output, where $k$ is a hyperparameter deciding the size of the generated output.
For the controlled setting, we select only the top sentences with the right category labels according to the control sequence. To do so, we employ an LSTM-CRF \cite{lample2016neural} tagger trained on the labeled meta-reviews to predict the labels of each input review sentence. Refer to Appendix \ref{append:classification} for more details of the tagger.




\begin{figure*}[t!]
    \begin{center}
    \resizebox{\textwidth}{!}{
    \includegraphics[width=\textwidth,height=30mm]
    {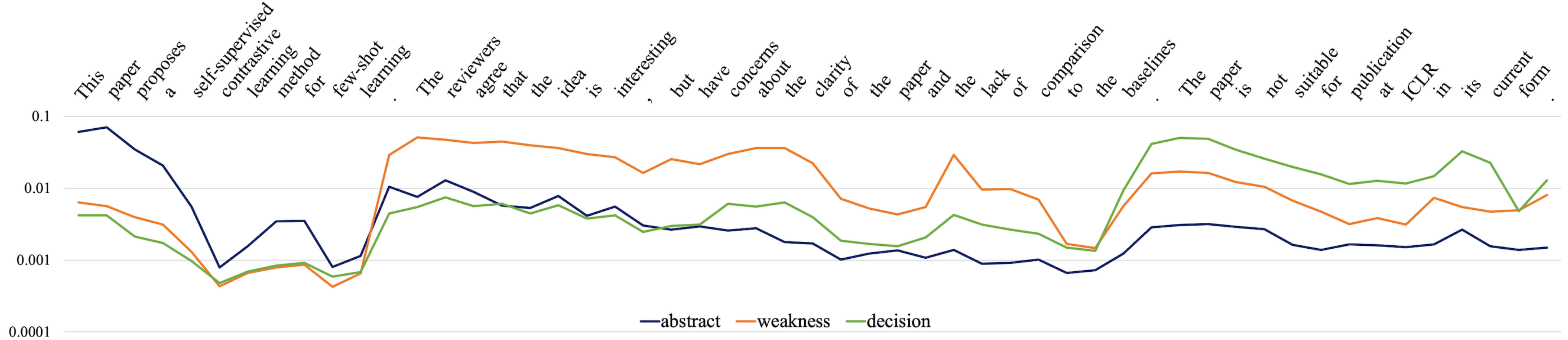}}
    \end{center}
    \caption{ Cross attention weights of each generated token towards the control tokens in logarithmic scale.
    }
    \label{fig:label_attn}
\end{figure*}

\paragraph{Generic Sentence Baselines.} 
Considering the nature of meta-reviews,
we could imagine some categories may have common phrases inflating the Rouge scores, such as ``This paper proposes ...'' for abstract, and ``I recommend acceptance.'' for decision, etc.
To examine such impact, we select sentences that are generic in each category and combine these sentences to generate outputs according to the control sequences.
For instance, if the control sequence is ``abstract | strength | decision'', we take the most generic sentences from the categories of ``abstract'', ``strength'' and ``decision'' respectively to form the output.
Specifically, we create two generic sentence baselines by obtaining generic sentences from the training data from either the meta-review references (i.e., target) or the input reviews (i.e., source), namely ``Target Generic'' and ``Source Generic''. 
Moreover, we also study such impact on the high-score and low-score submissions respectively, since an AC may write more succinct meta-reviews for clear-cut papers, as suggested by Figure \ref{fig:length_score}. See Appendix \ref{append:gen_baselines} for more details and results on generic sentence baselines.

\subsection{Experimental Setting}

To conduct text generation experiments, we preprocess our MReD dataset by filtering to ensure the selected meta-reviews have 20 to 400 words, as certain meta-review passages are extremely short or long. 
After preprocessing, we obtain 6,693 source-target pairs, for which we randomly split into train, validation, and test sets by a ratio of 8:1:1. 
We evaluate our generated outputs against the reference meta-reviews using the F$_1$ scores of ROUGE$_1$, ROUGE$_2$, and ROUGE$_L$ \cite{lin2004rouge}~\footnote{We use the Hugging Face Transformers' Rouge evaluation script, which has the field ``use\_stemmer'' enabled. We include the evaluation script in our code.}.

For the extractive and generic baselines, a key hyperparameter is the sentence number $k$.
Recall that under the \textit{sent-ctrl} setting, the control sequence length is the same as the sentence number of the target meta-review. Therefore, to conduct a fair comparison, we set the hyperparameter $k$ equal to the number of labels in the control sequence for both controlled and uncontrolled extractive baselines, and \textit{sent-ctrl} is used for all controlled extractive baselines. 
We also adopt the same $k$ for the generic baselines. 

For bart-large-cnn, we first load the pretrained model and then fine-tune it on MReD.
All experiments are conducted on single V100 GPUs,
using a batch size of 1 in order to fit the large pretrained model on a single GPU.
During fine-tuning, we set the hyperparameters of ``minimum\_target\_length'' to 20, and ``maximum\_target\_length'' to 400, according to our filter range on the meta-review lengths. 
Due to long inputs (see Table \ref{append:length-stats}), we experiment with different source truncation lengths of 1024, 2048, and 3072 tokens.
We cannot explore truncation length of more than 3072 tokens due to the limitation of GPU space.
Our learning rate is 5e-5, and we use Adam optimizer with momentum $\beta_1 = 0.9$, $\beta_2 = 0.999$ without any warm-up steps or weight decay. 
We set the seed to be 0, and train the model for 3 epochs with gradient accumulation step of 1.
For decoding, we use a beam size of 4 and length penalty of 2.

\begin{table}[t!]
	\centering
    \resizebox{0.8\columnwidth}{!}{
	    \setlength{\tabcolsep}{1mm}{
        \begin{tabular}{lccc}
        \toprule
        & R$_1$ & R$_2$ & R$_L$ \\
        \midrule
        Source Generic & 27.58 & 3.97 & 14.14 \\
        Target Generic & \textbf{27.98} & \textbf{5.52} & \textbf{15.01} \\
        \midrule
        MMR, \textit{unctrl} 
        & 31.43 & 5.45 & 16.31\\
        LexRank, \textit{unctrl} 
        & 31.74 & 6.67 & 16.71 \\
        TextRank, \textit{unctrl} 
        & 32.72 & \textbf{7.37} & 17.25\\
        MMR, \textit{sent-ctrl} & 32.37 & 6.28 & 17.58\\
        LexRank, \textit{sent-ctrl} & 32.60 & 6.66 & 17.48 \\
        TextRank, \textit{sent-ctrl} & \textbf{33.52} & 7.20 & \textbf{17.75}\\
        \midrule
        bart-large-cnn, \textit{unctrl}
        & 33.31 & 8.63 & 19.67 \\
        bart-large-cnn, \textit{sent-ctrl} \textcolor{white}{ccccc} & \textbf{38.73} & \textbf{10.82} & \textbf{23.05} \\
        bart-large-cnn, \textit{seg-ctrl} & 36.38 & 10.04 & 21.90 \\
        \bottomrule
        \end{tabular}}}
    \caption{Meta-review generation results on MReD.}
	\label{tab:generation}
\end{table}

\subsection{Main Results}
\label{section:results}
We show results in Table \ref{tab:generation}. Only the best settings of \textit{rate-concat} (
Section \ref{section:rev_comb}) and input truncation of 2048 tokens (Appendix \ref{append:more_result}) for bart-large-cnn are included.
Amongst the extractive baselines, TextRank performs the best in both \textit{unctrl} and \textit{sent-ctrl} settings.
Nevertheless, all controlled methods outperform their \textit{unctrl} settings (same for the Transformers).
This validates our intuition that structure-controlled generation is more suitable for user-subjective writings such as meta-reviews, because the model can better satisfy different structure requirements when supplied with the corresponding control sequences. 
On the other hand, for bart-large-cnn, \textit{sent-ctrl} is the best, followed by \textit{seg-ctrl}. 
This is most likely due to the former's more fine-grained sentence-level control that provides a clearer structure outline, as compared to the coarser segment-level control.

Moreover, bart-large-cnn far outperforms the extractive baselines, showing that the extraction-based methods are insufficient for MReD.
This also suggests that meta-review writings are different from the input reviews, therefore copying full review sentences to form meta-reviews doesn't work well.
This is also validated by the ``Target Generic'' baseline's consistent improvement over the ``Source Generic'' baseline,
which shows that generic sentences from meta-reviews can suit generation better than those in reviews.
Nevertheless, all Transformers results are still much better than the ``Target Generic'' sentence baseline, showing that despite generic phrases in some categories contributing to Rouge, the Transformers model is capable of capturing content-specific information for each input.

\subsection{Review Combination Results}
\label{section:rev_comb}
We also show uncontrolled generation results for different review combination methods in Table \ref{append:methods}, with source truncation of 2048. The \textit{longest-review} setting has the worst performance, thus validating that the review combination methods are necessary in order not to omit important information. 
\textit{Rate-concat} has the best overall performance, which is the setting we used for the main results. 
Nevertheless, it is not significantly better than \textit{merge}. 
It is also interesting to see that for \textit{merge}, providing additional rating information (\textit{rate-merge}) slightly worsens the performance.
We will leave the investigation of better review combination methods for future work.

\begin{table}[t!]
	\centering
	\resizebox{0.8\columnwidth}{!}{
	    \setlength{\tabcolsep}{1mm}{
        \begin{tabular}{lccc}
        \toprule
         & R$_1$ & R$_2$ & R$_L$ \\
        \midrule
        longest-review \textcolor{white}{cccccccccccccc} & 32.07 & 7.86 & 19.00 \\
        concat & 32.88 & 8.58 & 19.63 \\
        merge & 33.19 & \textbf{8.77} & 19.31 \\
        rate-concat & \textbf{33.31} & 8.63 & \textbf{19.67} \\
        rate-merge & 33.05 & 8.54 & 19.01 \\
        \bottomrule
        \end{tabular}}}
    \caption{Meta-review uncontrolled generation results for different review combination methods.}
	\label{append:methods}
\end{table}

\begin{table*}[t!]
	\centering
	\scalebox{0.7}{
	    \setlength{\tabcolsep}{1.5mm}{
        \begin{tabular}{p{0.5cm}@{~} p{3cm} p{18.5cm}}
        \toprule
        & \textbf{Gold Labels} & \textbf{Gold Meta-review} \\
        \midrule
        0 & abstract | weakness | decision & \textcolor{blue}{[}The paper presents a self-supervised model based on a contrastive autoencoder that can make use of a small training set for upstream multi-label/class tasks.\textcolor{blue}{]$\leftarrow$\textsc{abstract}}
        \textcolor{blue}{[}Reviewers have several concerns, including the lack of comparisons and justification for the setting, as well as the potentially narrow setting.\textcolor{blue}{]$\leftarrow$\textsc{weakness}}
        \textcolor{blue}{[}Overall, I found the paper to be borderline, the cons slightly greater than the pros, so I recommend to reject it.\textcolor{blue}{]$\leftarrow$\textsc{decision}} \\
        \toprule
        & \textbf{Control Sequence} & \textbf{Generated Passage} \\
        \midrule
        1 & abstract | weakness | decision &
        \textcolor{blue}{[}This paper proposes a self-supervised contrastive learning method for few-shot learning.\textcolor{blue}{]$\leftarrow$\textsc{abstract}}
        \textcolor{blue}{[}The reviewers agree that the idea is interesting, but have concerns about the clarity of the paper and the lack of comparison to the baselines.\textcolor{blue}{]$\leftarrow$\textsc{weakness}}
        \textcolor{blue}{[}The paper is not suitable for publication at ICLR in its current form.\textcolor{blue}{]$\leftarrow$\textsc{decision}} \\
        \midrule
        2 & abstract | abstract | weakness | decision & 
        \textcolor{blue}{[}This paper proposes a self-supervised contrastive learning method for few-shot learning.\textcolor{blue}{]$\leftarrow$\textsc{abstract}}
        \textcolor{blue}{[}The idea is to learn the label embeddings using in-task data, and then use those for training a contrastive autoencoder for multi-label classification on the long-tail problem.\textcolor{blue}{]$\leftarrow$\textsc{abstract}}
        \textcolor{blue}{[}The reviewers and AC note the following potential weaknesses: (1) limited novelty, (2) limited evaluation, and (3) lack of comparison to relevant baselines.\textcolor{blue}{]$\leftarrow$\textsc{weakness}} 
        \textcolor{blue}{[}AC agrees with the reviewers that the paper is not ready for publication in its current form.\textcolor{blue}{]$\leftarrow$\textsc{decision}} \\
        \midrule
        3 & strength | weakness | weakness | rebuttal process | suggestion | decision &
        \textcolor{blue}{[}Pros: The paper is well-motivated and the idea is simple and straightforward.\textcolor{blue}{]$\leftarrow$\textsc{strength}}
        \textcolor{blue}{[}However, the reviewers have raised several concerns about the evaluation and the clarity of the paper.\textcolor{blue}{]$\leftarrow$\textsc{weakness}}
        \textcolor{blue}{[}In particular, the lack of comparison to the baselines is a major concern.\textcolor{blue}{]$\leftarrow$\textsc{weakness}}
        \textcolor{blue}{[}The authors have addressed some of these concerns in the rebuttal, but the reviewers are still not convinced about the significance of the results. \textcolor{blue}{]$\leftarrow$\textsc{rebuttal process}}
        \textcolor{blue}{[}The paper would be much stronger if the authors could compare their method to more baselines for zero-shot learning, such as matching networks and Siamese networks.\textcolor{blue}{]$\leftarrow$\textsc{suggestion}}
        \textcolor{blue}{[}Overall, the paper is not ready for publication at ICLR.\textcolor{blue}{]$\leftarrow$\textsc{decision}} \\
        \bottomrule
        \end{tabular}}}
    \caption{Varied generation outputs by giving different control sequences.}
	\label{tab:control_gen_examples}
\end{table*}

\begin{table}[ht]
	\centering
	\resizebox{\columnwidth}{!}{
	    \setlength{\tabcolsep}{0mm}{
            \begin{tabular}{p{2.8cm}@{~} p{5mm}@{~} @{~}p{6.5cm}}
            \toprule
            \textbf{Generated \newline Content} & & \textbf{Attention Attribution}\newline   \includegraphics[width=6cm,right]{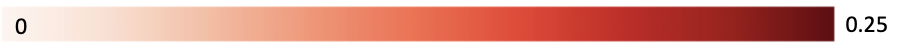} \\
            \midrule
            \scriptsize \textbf{Sent 1 (abstract): } \newline This paper proposes a self-supervised contrastive learning method for few-shot learning. 
            & & \includegraphics[align=t,width=6.5cm]{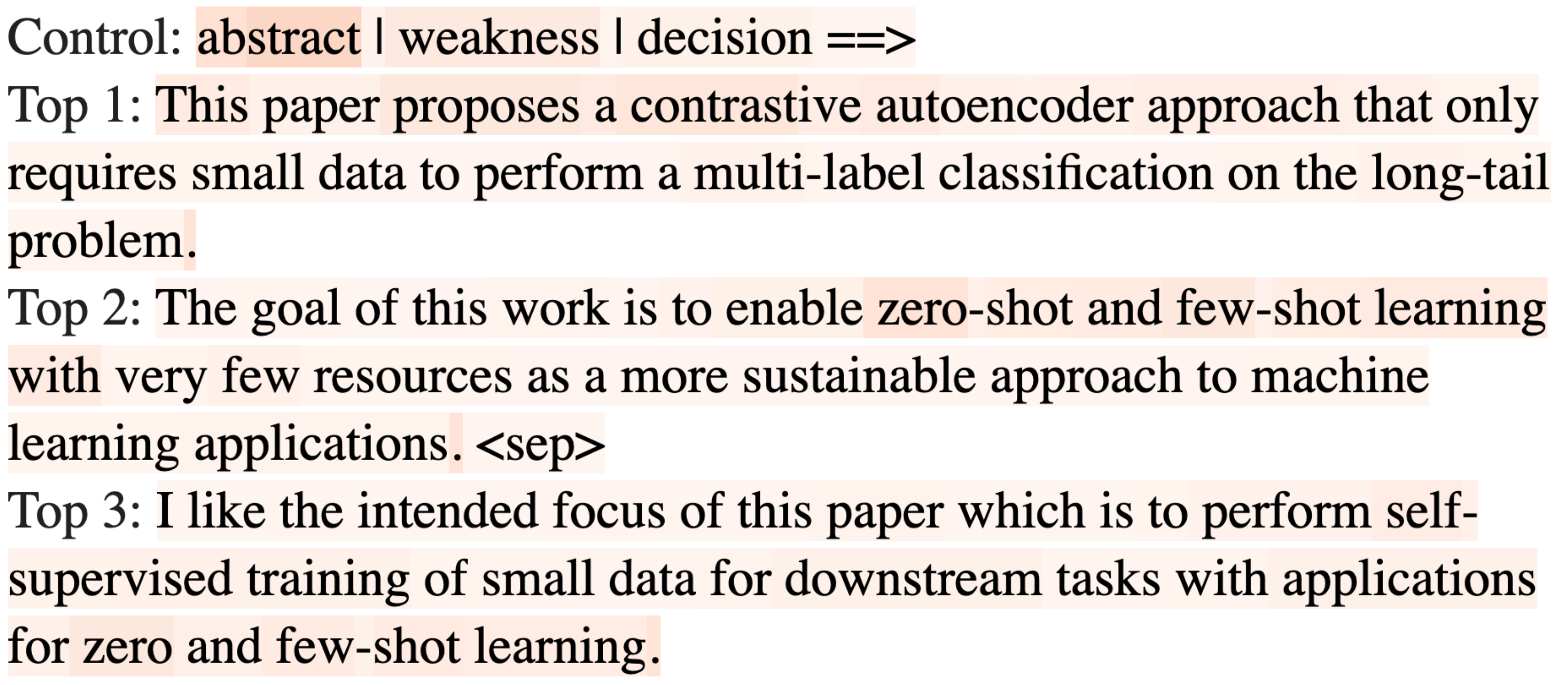} \newline \\
            \hline
            \scriptsize
            \textbf{Sent 2 (weakness): } \newline
            The reviewers agree that the idea is interesting, but have concerns about the clarity of the paper and the lack of comparison to the baselines. 
            & & \includegraphics[align=t,width=6.5cm]{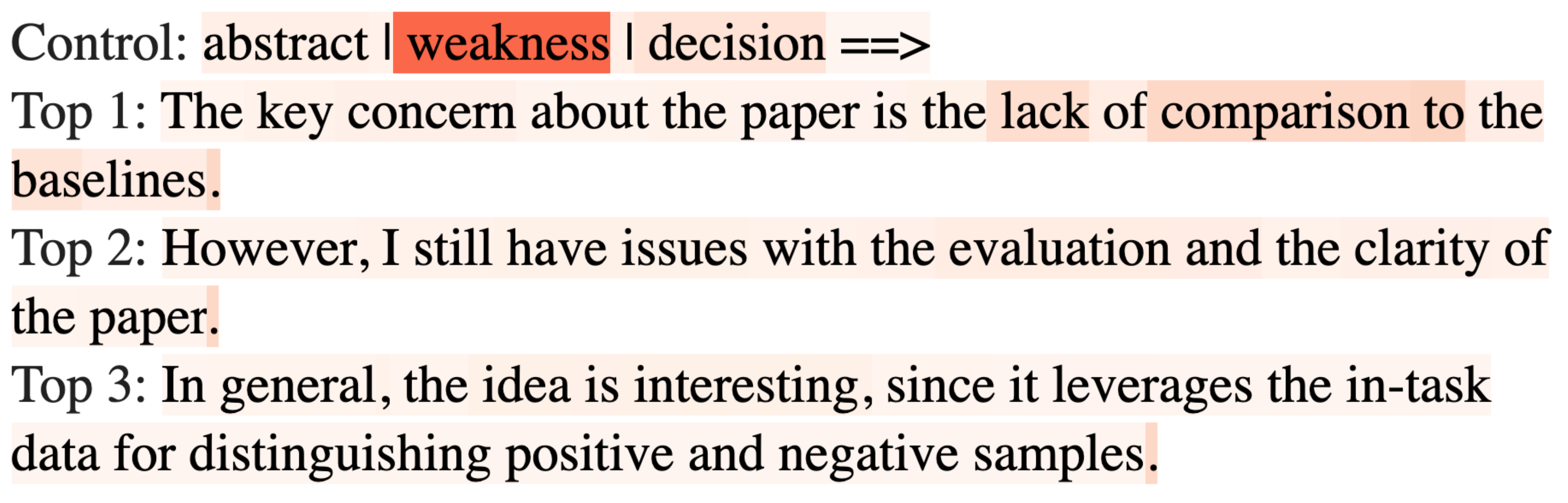} \newline \\
            \hline
            \scriptsize
            \textbf{Sent 3 (decision): } \newline
            The paper is not suitable for publication at ICLR in its current form.
            & & \includegraphics[align=t,width=6.5cm]{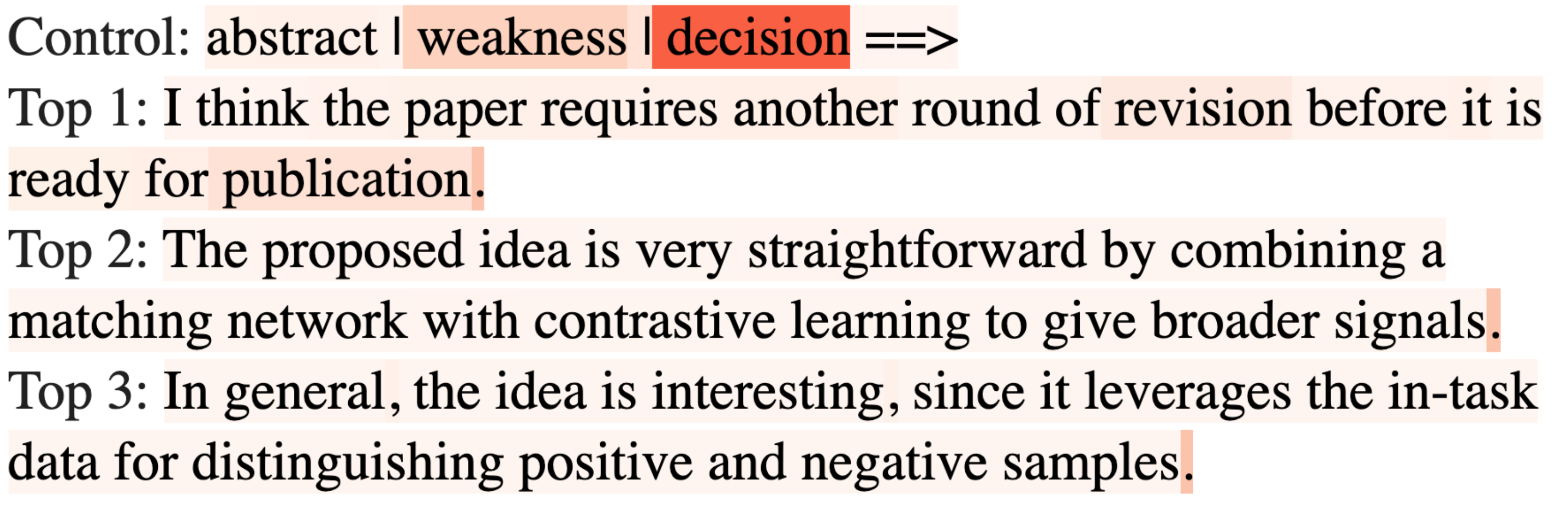} \newline \\
            \bottomrule
            \end{tabular}
        }
    }
    \caption{Attention analysis for each output sentence.}  
	\label{tab:attn_attr}
\end{table}


\subsection{Case Study}
\label{case}

We study some cases for a better understanding of the structure-controllable generation. 

\paragraph{Identify the control label for each sentence.}
We first evaluate whether the model is able to attend to the correct control label during generation.
For each generation step, we obtain the cross attention weights from the decoder's output token towards the control labels, and plot them in Figure \ref{fig:label_attn}.
The given control sequence is ``abstract | weakness | decision''. 
When generating each sentence, we can see that the attention weights of the corresponding control token are the highest, which demonstrates that our model can effectively pay attention to the correct control label and thus generate the content complying with the intent.

\paragraph{Extract information from the input sentences.}
To understand what information the model attends to when generating each sentence, 
we aggregate the cross attention weights to obtain the attention scores from each generated sentence towards all input sentences (Appendix \ref{append:attn-aggr-method}).
Then, we select the top 3 input sentences with the highest attention scores for each generated sentence, and visualize the normalized attention weights on all tokens in the selected sentences and the control sequence in Table \ref{tab:attn_attr}.
As shown, the model can correctly extract relevant information from the source sentences.
For example, it identifies important phrases such as ``interesting'', ``clarity'' and ``lack of comparison to baselines'' when generating ``Sent 2''.

\paragraph{Generate varied outputs given different control sequences.}
To further investigate the effectiveness of the control sequence, we change the control sequence of the above example and re-generate the meta-reviews given the same input reviews.
In Table \ref{tab:control_gen_examples}, we first show the gold meta-review and the model output using the original control sequence in Row 0 and Row 1, 
and then show the model outputs with alternative control sequences in Row 2 and Row 3.
From the outputs, we can see that indeed each generated sentence corresponds to its control label well.
In Row 2, we add an additional control label in the sequence and by repeating the ``abstract'' label, the generator can further elaborate more  details of the studied method.
This is one key advantage of our \textit{sent-ctrl} compared to the \textit{seg-ctrl}, which allows the control of length and the level of the generation details.
In Row 3, a very comprehensive control sequence is specified. We can see that the output meta-review is quite fluent and polite to reject the borderline paper. See Appendix \ref{append:case_study} for more examples.

\subsection{Human Evaluation}


In addition to the Rouge evaluation, we ask 3 human judges to manually assess the generation quality of the bart-large-cnn model trained under different control methods from Table \ref{tab:generation} on 100 random test instances.
For each test instance, we provide the judges with the input reviews and randomly ordered generations from different models, and ask them to individually evaluate the generations based on the following criteria:
(1) \textit{Fluency}: is the generation fluent, grammatical, and without unnecessary repetitions?
(2) \textit{Content Relevance}: does the generation reflect the review content well, or does it produce general but trivial sentences? 
(3) \textit{Structure Similarity}: how close does the generation structure resemble the gold structure (i.e., the control sequence)?
(4) \textit{Decision Correctness}: 
does the generation correctly predicts the gold human decision? 
We grade fluency and content relevance on a scale of 1 to 5, whereas structure similarity and decision correctness are calculated from 0 to 1 (Appendix \ref{append:human_eval}).
For structure similarity, because \textit{sent-ctrl} and \textit{seg-ctrl} have different control sequences, we evaluate the two models on sentence-level (sent) and segment-level (seg) structures respectively, and provide both evaluations for \textit{unctrl}.

As shown in Table \ref{tab:human_eval}, both \textit{sent-ctrl} and \textit{seg-ctrl} models show significant improvements on the generation structure over the uncontrolled baseline, which affirms the effectiveness of our proposed methods for structure-controllable generation.
\textit{Sent-ctrl} also has better fluency and decision correctness, suggesting that having a better output structure can benefit readability and decision generation.
For the content relevance, the scores of all methods are reasonably good, and significance tests cannot prove any best model ($p>0.08$).
Nevertheless, it is possible that the looser control a method applies, the better relevance score it achieves. It is because a tighter control narrows the content that a model can use from the reviews.

\begin{table}[t!]
    \footnotesize
	\centering
    \resizebox{0.85\columnwidth}{!}{
	    \setlength{\tabcolsep}{1mm}{
        \begin{tabular}{lccc}
        \toprule
        & Unctrl & Sent-ctrl & Seg-ctrl \\
        \midrule
        Fluency & 4.145 & \textbf{4.630*} & 4.090 \\
        Content Relevance & \textbf{4.585} & 4.335 & 4.410 \\
        Structure Similarity (sent) & 0.298 & \textbf{0.706*} & - \\
        Structure Similarity (seg) & 0.363 & - & \textbf{0.623*} \\
        Decision Correctness & 0.685 & \textbf{0.830*} & 0.695 \\
        \bottomrule
        \end{tabular}}}
    \caption{Human evaluation. * indicates the ratings of corresponding models significantly (by Welch’s t-test) outperform the \textit{unctrl}: $p < 0.01$ for decision correctness, $p < 0.0001$ for fluency and structure similarity.
    }
	\label{tab:human_eval}
\end{table}


\section{Related Work}
To facilitate the study of text summarization, earlier datasets are mostly in the news domain with relatively short input passages, such as NYT \cite{sandhaus2008new}, Gigaword \cite{napoles2012annotated}, CNN/Daily Mail \cite{hermann2015teaching}, {N\textsc{ewsroom}} \cite{grusky2018newsroom} and XS\textsc{um} \cite{narayan2018don}. 
Datasets for long documents include \citet{sharma2019bigpatent}, \citet{cohan2018discourse}, and \citet{fisas2016multi}.
In this paper, we explore text summarization in a new domain (i.e., the peer review domain) and provide a new dataset, i.e., MReD. Moreover, MReD's reference summaries (i.e., meta-reviews) are fully annotated and thus allow us to propose a new task, namely, structure-controllable text generation.

Researchers recently explore the peer review domain data for a few tasks, such as PeerRead \cite{kang2018dataset} for paper decision predictions, AMPERE \cite{hua2019argument} for proposition classification in reviews, and RR \cite{cheng2020argument} for paired-argument extraction from review-rebuttal pairs. 
Additionally, a meta-review dataset is introduced by \citet{bhatia2020metagen} without any annotation.
Our work is the first fully-annotated dataset in this domain for the structure-controllable generation task.
There are also some datasets and annotation schemes on research articles \cite{teufel1999annotation, liakata2010corpora, lauscher2018arguminsci}, which differ in nature from the peer review domain and cannot be easily transferred to our task.


A wide range of control perspectives has been explored in controllable generation, including style control (e.g., sentiments \cite{duan2020pre}, politeness \cite{madaan2020politeness}, formality \cite{wang2019harnessing}, domains \cite{takeno2017controlling} and persona \cite{zhang2018personalizing}) and content control (e.g., length \cite{duan2020pre}, entities \cite{fan2018controllable}, and keywords \cite{tang2019target}).
Our structure-controlled generation differs from these works as we control the high-level output structure, rather than the specific styles or the surface details of which keywords to include in the generated output. 
Our task also differs from content planning \cite{reiter1997building, shao2019long, hua2019sentence}, which involves explicitly selecting and arranging the input content.
Instead, we provide the model with the high-level control labels, and let the model decide on its own the relevant styles and contents.

\section{Conclusions}

This paper introduces a fully-annotated text generation dataset MReD in a new domain, i.e., the meta-reviews in the peer review system, and provides thorough data analysis to better understand the data characteristics.
With such rich annotations, we propose simple yet effective methods for structure-controllable text generation.
Extensive experimental results are presented as baselines for future study and thorough result analysis is conducted to shed light on the control mechanisms.

\section{Ethical Concerns}
We have obtained approval from ICLR organizers to use the data collected from ICLR 2018-2021 on OpenReview.

\bibliography{anthology,custom}
\bibliographystyle{acl_natbib}

\appendix

\clearpage

\section{Data Annotation}
\begin{table}[t!]
	\centering
	\resizebox{\columnwidth}{!}{
	    \setlength{\tabcolsep}{1mm}{
        \begin{tabular}{p{2.8cm}p{9cm}}
        \toprule
        Categories & Examples \\
        \midrule
        \textbf{abstract} & ``The paper presents/explores/describes/addresses/proposes ...'' \\
        \midrule
        \textbf{strength}  & ``The reviewers found the paper interesting.'' ``The method and justification are clear.'' ``The quantitative results are promising.'' \\
        \midrule
        \textbf{weakness} & ``The paper is somewhat incremental ...'' ``... claims are confusing'' ``The main concern is ...'' ``... unfair experimental comparisons ...'' \\
        \midrule
        \textbf{rating summary} & ``R1 recommends Accept.'' ``All four reviewers ultimately recommended acceptance.'' ``Reviews were somewhat mixed, but also with mixed confidence scores.'' \\
        \midrule
        \textbf{ac disagreement} & ``The area chair considers the remaining concerns by Reviewer 3 as invalid.'' ``I do not agree with the criticism about ...'' ``I disagree with the second point ...'' \\
        \midrule
        \textbf{rebuttal process} & ``The authors have made various improvements to the paper'' ``... remained after the author rebuttal ...'' ``Authors provided convincing feedbacks on this key point.'' \\
        \midrule
        \textbf{suggestion} & ``... more analysis ...'' ``The authors are advised to take into account the issues about ...'' \\
        \midrule
        \textbf{decision} & ``The paper is recommended as a poster presentation.'' ``AC recommends Reject.'' ``I recommend rejection.'' \\
        \midrule
        \textbf{miscellaneous} & ``Thank you for submitting you paper to ICLR.'' ``I've summarized the pros and cons of the reviews below.'' \\
        \bottomrule
        \end{tabular}}}
    \caption{Category examples of meta-review sentences.
    }
	\label{append:category}
	\vspace{-2mm}
\end{table}

\subsection{Category definitions}
\label{append:category_definition}
We show category examples in Table \ref{append:category}.

\subsection{Additional annotation rules}
\label{append:additional_rule}
The additional rules for annotation are as follows:
First, instead of only labeling the individual sentences per se, the annotators are given a complete paragraph of meta-review to label the sentences with context information.
For example, if the area chair writes a sentence providing some extra background knowledge in the discussion of the weakness of the submission, even though that sentence itself can be considered as ``misc'', it should still be labeled as ``weakness'' to be consistent in context.

Second, not every sentence can be strictly classified into a single category. When a sentence contains information from multiple categories, the annotators should consider its main point and primary purpose. 
One example is: ``Although the paper discusses an interesting topic and contains potentially interesting idea, its novelty is limited.''
Although the first half of the sentence discusses the strength of the submission, the primary purpose of this sentence is to point out its weakness, and therefore it should be labeled as weakness.

Furthermore, there are still some cases where the main point of the sentence is hard to differentiate from multiple categories.
We then define a priority order of these 9 categories according to the importance of each category for annotators to follow: decision $>$ rating summary $>$ strength $\stackrel{?}{=}$ weakness $>$ ac disagreement $>$ rebuttal process $>$ abstract  $>$ suggestion $>$ miscellaneous.
We use the sign ``$\stackrel{?}{=}$'' because there are some rare cases where a sentence contains both ``strength'' and ``weakness'' while there is no  obvious emphasis on either, and it is hard to tell whether ``strength'' should have a priority over ``weakness'' or the other way round.
We then label this sentence based on the final decision:
if this submission is accepted, we label the sentence as ``strength'', and vice versa.

\section{Data Analysis}
\subsection{Borderline papers}
\label{append:borderline_distri}

We further analyze the category distribution in borderline papers.
As shown in Table \ref{append:borderline}, for submissions within the score range of [4.5,6), there are 713 accepted submissions and 2,588 rejected submissions.
One clear difference is the percentage of ``strength'' and ``weakness''.
Another difference is the percentage of ``ac disagreement'', where the accepted papers have four times the value than rejected ones.
This suggests that for the accepted borderline papers, the area chair tends to share different opinions with reviewers, and thus deciding to accept the borderline submissions.


\begin{table}[t!]
    \footnotesize
	\centering
    \resizebox{\columnwidth}{!}{
	    \setlength{\tabcolsep}{1mm}{
        \begin{tabular}{lcc}
        \toprule
        & Accept & Reject \\
        \midrule
        abstract & 23.8\% & 18.1\% \\
        strength & 18.1\% & 9.3\% \\
        weakness & 13.5\% & 34.3\% \\
        rating summary & 6.3\% & 4.1\% \\
        ac disagreement &  2.2\% & 0.5\% \\
        rebuttal process \textcolor{white}{ccccccccccccccccccc} & 13.2\% & 11.0\% \\
        suggestion & 7.7\% & 8.2\% \\
        decision & 9.2\% & 8.1\% \\
        miscellaneous & 6.2\% & 6.4\% \\
        \bottomrule
        \end{tabular}}}
    \caption{Category distribution of borderline submissions (average score in the range of [4.5,6) breakdown by final decision.
    }
	\label{append:borderline}
	\vspace{-3mm}
\end{table}

\begin{table}[t!]
    \footnotesize
	\centering
    \resizebox{\columnwidth}{!}{
	    \setlength{\tabcolsep}{1mm}{
        \begin{tabular}{lccc|ccc}
        \toprule
        & \multicolumn{3}{c}{Accept} & \multicolumn{3}{c}{Reject}\\
        & Low & Med & High & Low & Med & High \\
        \midrule
        abstract & \textbf{79} & 75 & 74 & 69 & 69 & 74 \\
        strength & 64 & \textbf{71} & 70 & 26 & 43 & 50 \\
        weakness & 49 & 44 & 32 & 79 & 84 & \textbf{88} \\
        rating summary & 25 & \textbf{33} & 32 & 29 & 25 & 24 \\
        ac disagreement & 1 & \textbf{6} & 2 & 1 & 2 & 3\\
        rebuttal process \textcolor{white}{ccccccccc} & \textbf{52} & 47 & 37 & 35 & 39 & 39  \\
        suggestion & 29 & 26 & 23 & 23 & 32 & \textbf{38} \\
        decision & \textbf{56} & 53 & 46 & 53 & 53 & \textbf{56} \\
        miscellaneous & 19 & 19 & 14 & 24 & 35 & \textbf{45} \\
        \bottomrule
        \end{tabular}}}
    \vspace{-2mm}
    \caption{Occurrence of different categories for accepted and rejected papers, breakdown by average scores. Low for scores $\leq$ 5.5, high for scores $\geq$ 6.5, and med for borderline scores in between.
    }
	\label{append:accept_reject}
	\vspace{-3mm}
\end{table}

\subsection{Percentage of each category for accepted and rejected papers across score ranges}
\label{append:accept_reject_breakdown}

We further analyze the occurrence of each category for accepted papers and rejected papers separately across different score ranges, as shown in Table \ref{append:accept_reject}. 
For accepted papers, as the score increases, the percentage of meta-reviews having ``weakness'' and ``suggestion'' drops because the high-score submissions are more likely to be accepted.
Even the percentage of ``decision'' drops following the same trend.
In addition, the proportion of meta-reviews having ``rebuttal process'' is larger for submissions with lower scores.
This suggests that the rebuttal process plays an important role in the peer review process, especially in helping the borderline papers to be accepted.

On the other hand, for rejected papers, the percentage of meta-reviews having ``strength'' increases as the average score increases.
This coincides with our common sense that the submissions receiving higher scores tend to have more strengths.
One interesting finding here is that the percentage of ``weakness'' and ``suggestion'' also increases as the average rating score increases.
This may be due to two main reasons.
First, to reject a submission with higher scores, the area chair has to explain the weakness with more details and provide more suggestions for authors to further improve their submissions.
Second, compared to the percentage of ``strength'', ``weakness'' definitely has a larger percentage within any range of rating scores.
The difference in the percentage of ``strength'' and ``weakness'' is intuitively different between the accepted papers and the rejected papers.

\section{Experiments}
\subsection{Additional transformers models}
\label{append:pretrained-models}
\begin{table}[t!]
	\centering
	\scalebox{0.85}{
	    \setlength{\tabcolsep}{1mm}{
            \begin{tabular}{lccc}
            \toprule
            Pretrained Model & R$_1$ & R$_2$ & R$_L$ \\
            \midrule
            \textbf{Uncontrolled Generation} & & &\\
            facebook/bart-large-cnn* \textcolor{white}{cccccccccc} & \textbf{33.20} & \textbf{8.55} & 19.62 \\
            facebook/bart-large & 28.86 & 6.20 & 19.02 \\
            t5-large & 30.75 & 8.44 & \textbf{20.23} \\
            google/pegasus-cnn\_dailymail & 28.76 & 6.37 & 16.79 \\
            \midrule
            \textbf{Controlled Generation, \textit{sent-ctrl}} & & & \\
            facebook/bart-large-cnn* \textcolor{white}{cccccccccc} & \textbf{38.39} & 10.60 & 22.86 \\
            facebook/bart-large & 38.05 & \textbf{10.66} & 23.39 \\
            t5-large & 35.90 & 10.18 & \textbf{23.92} \\
            google/pegasus-cnn\_dailymail & 33.48 & 8.68 & 21.03 \\
            \bottomrule
            \end{tabular}
        }
    }
    \caption{Results of other common Transformers summarization models using source truncation of 1024. * represents our selected model in the main paper. }
	\label{tab:pretrained_models}
	\vspace{-4mm}
\end{table}

\begin{table*}[t!]
	\centering
	\resizebox{\textwidth}{!}{
	    \setlength{\tabcolsep}{1.2mm}{
        \begin{tabular}{lcc|cccccccccc}
        \toprule
         & Micro F$_1$ & Macro F$_1$ & abstract & strength & weakness & rating & ACdisagree & rebuttal & suggestion & decision & misc \\
        \midrule
        BERT-base-cased + CRF & 85.27 & 76.71 & \textbf{94.58} & 86.12 & 86.21 & 85.21 & 30.77 & 73.80 & 73.89 & 91.30 & 68.49 \\
        BERT-large-cased + CRF & 84.68 & 77.84 & 93.93 & \textbf{86.71} & 84.36 & 84.07 & 40.00 & 72.60 & 74.35 & 91.60 & \textbf{72.96} \\
        RoBERTa-base + CRF &\textbf{85.83} & \textbf{79.99} & 94.47 & 86.43 & 86.73 & 84.56 & \textbf{54.84} & \textbf{74.44} & 72.79 & \textbf{93.08} & 72.54 \\
        RoBERTa-large + CRF & 85.72 & 79.34 & 94.42 & 85.61 & \textbf{87.09} & \textbf{85.40} & 50.00 & 73.97 & \textbf{75.63} & 90.93 & 71.00 \\
        \bottomrule
        \end{tabular}}}
    \caption{MReD sentence classification results.}
	\label{tab:discourse}
	\vspace{-3mm}
\end{table*}

We provide baselines of uncontrolled generation and controlled generation on MReD using other common Transformer pretrained models in Table \ref{tab:pretrained_models}. Note that due to limited GPU space, we cannot fit 2048 input tokens for T5. Thus, for fair comparison, all results shown are from source truncation of 1024.

\subsection{Tagger for source sentences}
\label{append:classification}

To obtain labels on source input, we train a tagger based on the human-annotated meta-reviews, then use it to predict labels on the input sentences.
Specifically, we define the task as a sequence labeling problem and apply the long short-term memory (LSTM) \cite{hochreiter1997long} networks with a conditional random field (CRF) \cite{lafferty2001conditional} (i.e., LSTM-CRF \cite{lample2016neural}) model on the annotated MReD dataset.
The same data split as the meta-review generation task is used.
We adopt the standard \textsc{IOBES} tagging scheme \cite{ramshaw1995text,ratinov2009design}, and fine-tune BERT \cite{kenton2019bert} and RoBERTa \cite{liu2019roberta} models in Hugging Face.
All models are trained for 30 epochs with an early stop of 20, and each epoch takes about 30 minutes.
We select the best model parameters based on the best micro F$_1$ score on the development set and apply it to the test set for evaluation.
All models are run with single V100 GPUs.
We use Adam \cite{kingma2014adam} with an initial learning rate of 2e-5.

We report the F$_1$ scores for each category as well as the overall micro F$_1$ and macro F$_1$ scores in Table \ref{tab:discourse}. 
Micro F1 is the overall accuracy regardless of the categories, whereas macro F1 is an average of per category accuracy evaluation. 
Since some of the category labels (eg. ``ac disagreement'') are very rare, their classification accuracy is low. 
Overall, micro F1 is a more important metric since it suggests general performance. 
The results stand proof that the majority of the categories have their own characteristics that can be identified from other categories.
RoBERTa\-base is the best performing model, therefore we use this model to predict review sentence labels.

\subsection{Generic sentence baselines}
\label{append:gen_baselines}

\begin{table}[t!]
	\centering
    \resizebox{\columnwidth}{!}{
	    \setlength{\tabcolsep}{1mm}{
        \begin{tabular}{lccc}
        \toprule
        & R$_1$ & R$_2$ & R$_L$ \\
        \midrule
        Source Generic & \textbf{27.58} & 3.97 & 14.14 \\
        Source High Score \textcolor{white}{cccccccccccccccc} & 26.95 & \textbf{4.38} & \textbf{15.18} \\
        Source Low Score & 25.82 & 4.14 & 14.40 \\
        \midrule
        Target Generic & 27.98 & 5.52 & 15.01 \\
        Target High Score & 31.10 & 5.76 & 16.82 \\
        Target Low Score & \textbf{32.04} & \textbf{7.21} & \textbf{19.09} \\
        \bottomrule
        \end{tabular}}}
    \caption{MReD generic sentence baseline results on various score subsets.}
	\label{append:generic_results}
\end{table}

Besides the baselines of ``Source Generic'' and ``Target Generic'', we explore subsets of papers with high scores (average reviewers' rating $\geqslant7$) or low scores (average reviewers' rating $\leqslant3$) to obtain 4 additional generic baselines: ``Source High Score'', ``Source Low Score'', ``Target High Score'', ``Target Low Score''. 

We use ``Target High Score'' as an example to explain how we obtain the generic sentences:
From the training subset of high score papers, We first separate all meta-review sentences into the corresponding label categories, obtaining a total of 9 groups of sentences.
Then, we re-arrange the sentences in each group using TextRank (our best extractive model). 
Since TextRank ranks the input sentences based on each sentence's content connection with others,
sentences with higher rankings are also more general in the sense that they have more shared content with others.

After obtaining the generic sentence sets, we can create baseline generations using the \textit{sent-ctrl} sequence on the corresponding high score paper test data.
We avoid using the same sentence twice inside the same generation, so if the same label appears multiple times in a control sequence, we will use the same number of generic sentences for that category down the ranking order.

All generic sentence baselines can be obtained in a similarly procedure as outlined above, and we show results in Table \ref{append:generic_results}.
Both ``Target High Score'' and ``Target Low Score'' perform much better than the ``Target Genric'' baseline, suggesting that papers with very high or low scores tend to have more typical patterns in their meta-reviews. Nevertheless, the pattern is less evident in the source (reviews) baselines. 

\begin{table*}[t!]
	\centering
	\scalebox{0.7}{
	    \setlength{\tabcolsep}{1.5mm}{
        \begin{tabular}{p{0.5cm}@{~} p{3cm} p{18.5cm}}
        \toprule
        & \textbf{Gold Labels} & \textbf{Gold} \\
        \midrule
        0 & abstract | \newline rating summary &  
        \textcolor{blue}{[}this work adapts cycle GAN to the problem of decipherment with some success.\textcolor{blue}{]$\leftarrow$\textsc{abstract}} 
        \textcolor{blue}{[}it's still an early result, but all the reviewers have found it to be interesting and worthwhile for publication.\textcolor{blue}{]$\leftarrow$\textsc{rating summary}}\\
        \toprule
        & \textbf{Control Sequence} & \textbf{Generated Passage} \\
        \midrule
        1 & abstract | \newline rating summary &  
        \textcolor{blue}{[}This paper proposes a method for using cycle GANs to decipher text encrypted with historical ciphers.\textcolor{blue}{]$\leftarrow$\textsc{abstract}} \newline 
        \textcolor{blue}{[}The reviewers were unanimous in their vote to accept.\textcolor{blue}{]$\leftarrow$\textsc{rating summary}} \\
        \midrule
        2 & abstract | abstract | \newline rating summary & \textcolor{blue}{[}This paper proposes a method for using cycle GANs to decipher text encrypted with historical ciphers.\textcolor{blue}{]$\leftarrow$\textsc{abstract}} 
        \textcolor{blue}{[}The proposed method is based on replacing the log loss of the GAN with a quadratic loss and regularizing the Jacobian of the discriminator.\textcolor{blue}{]$\leftarrow$\textsc{abstract}} \newline 
        \textcolor{blue}{[}The reviewers were unanimous in their vote to accept.\textcolor{blue}{]$\leftarrow$\textsc{rating summary}} \\
        \midrule
        3 & strength | decision & 
        \textcolor{blue}{[}The reviewers agree that the paper is well-written, the problem is interesting, and the proposed method is novel.\textcolor{blue}{]$\leftarrow$\textsc{strength}} 
        \textcolor{blue}{[}The paper should be accepted.\textcolor{blue}{]$\leftarrow$\textsc{decision}} \\
        \midrule
        4 & misc | abstract | \newline strength | decision & 
        \textcolor{blue}{[}Thanks for your submission to ICLR.\textcolor{blue}{]$\leftarrow$\textsc{misc}} \newline 
        \textcolor{blue}{[}This paper proposes a method for using cycle GANs to decipher text encrypted with historical ciphers.\textcolor{blue}{]$\leftarrow$\textsc{abstract}} 
        \textcolor{blue}{[}This is an interesting and fairly original problem, and the proposed method is shown to be very effective.\textcolor{blue}{]$\leftarrow$\textsc{strength}} 
        \textcolor{blue}{[}All reviewers agree that the paper is well written, and I'm happy to recommend acceptance.\textcolor{blue}{]$\leftarrow$\textsc{decision}} \\
        \midrule
        5 & weakness | \newline rebuttal process & 
        \textcolor{blue}{[} The reviewers raised a number of concerns including the correctness of the proof, the lack of a simple baseline, and the presentation of the paper.\textcolor{blue}{]$\leftarrow$\textsc{weakness}}
        \textcolor{blue}{[}The authors' rebuttal addressed some of these concerns, but not to the degree that the reviewers felt it should be.\textcolor{blue}{]$\leftarrow$\textsc{rebuttal process}}
        \\
        \bottomrule
        \end{tabular}}}
    \caption{Generation examples of alternative control sequences on the same review inputs using the \textit{sent-ctrl} method.}
	\label{tab:append_control_gen_examples}
\end{table*}

\label{append:setup}
\begin{table}[t!]
	\centering
	\scalebox{0.85}{
	    \setlength{\tabcolsep}{1mm}{
            \begin{tabular}{lccc}
            \toprule
            Data Split \textcolor{white}{cccccccccccccccccccccccc} & max & med & avg \\
            \midrule
            train & 7276 & 1482 & 1368 \\
            validation & 3762 & 1427 & 1352 \\
            test & 5144 & 1454 & 1352 \\
            \bottomrule
            \end{tabular}
        }
    }
    \vspace{-2mm}
    \caption{Source length statistics on all data splits. Max for maximum source length, med for median source length, and avg for average source length.}
	\label{append:length-stats}
\end{table}

\subsection{Ablation on truncation length}
\label{append:more_result}

\begin{table}[t!]
	\centering
	\scalebox{0.85}{
	    \setlength{\tabcolsep}{1mm}{
        \begin{tabular}{lccc}
        \toprule
        length & R$_1$ & R$_2$ & R$_L$ \\
        \midrule
        1024 \textcolor{white}{cccccccccccccccccccccccccc} & 38.39 & 10.60 & 22.86 \\
        2048 \textcolor{white}{cccccccccccccccccccccccccc} & \textbf{38.73} & \textbf{10.82} & \textbf{23.05} \\
        3072 \textcolor{white}{cccccccccccccccccccccccccc} & 38.30 & 10.34 & 22.57 \\
        \bottomrule
        \end{tabular}}}
    \caption{Meta-review \textit{sent-ctrl} generation results of different source truncation lengths. 
    }
	\label{append:src_trunc_results}
	\vspace{-3mm}
\end{table}

By default, the Transformers truncate the source to 1024 tokens. We further investigate the performance of different source truncation lengths under the setting of \textit{rate-concat}.
As shown in Table \ref{append:src_trunc_results}, truncating the source to 2048 tokens consistently achieves the best performance.

\subsection{Attention aggregation method}
\label{append:attn-aggr-method}
During generation, we can obtain the attention weights of each output token towards all input tokens.
Specifically, we average all decoder layers' cross attention weights for the same output token generated at each decoding step.
We then calculate an attention value for that output token on each input sentence, by aggregating the token's attention weights on the list of input tokens that belong to the same sentence by max pooling.
Finally, we can calculate an output-sentence-to-input-sentence attention score, by adding up these attention values for the output tokens that belong to the same sentence.

Common attention aggregation methods include summation, average-pooling, and max-pooling.
We use max-pooling to aggregate attention for same-sentence input tokens, because summation unfairly gives high attention scores to excessively long sentences due to attention weight accumulation, whereas average-pooling disfavors long sentences containing a few relevant phrases by averaging the weights out.
With max-pooling, we can correctly identify sentences with spiked attention at important phrases, regardless of sentence lengths.
For attention aggregation on the same-sentence output tokens, summation is used and can be viewed as allowing each output token to vote an attention score on all input sentences, so that the input sentence receiving the highest total score is the most relevant.
We conduct trial runs of all aggregation methods on input tokens with summation for output-token aggregation for multiple generation examples, and indeed max-pooling outperforms the other two by identifying more relevant input sentences with the generated sentence.

Once we have the attention scores, we can attribute the generation of each output sentence to a few topmost relevant input sentences. Then, we can draw a color map of the input tokens in the selected sentences based on their relative attention weights. 

\subsection{Structure-controlled generation examples} 
\label{append:case_study}
We show examples of the generation results using alternative control sequences on another submission in Table \ref{tab:append_control_gen_examples}.
We can see the effectiveness of controlling the output structure using our proposed method.

\subsection{Human evaluation}
\label{append:human_eval}
For structure similarity, we instruct the judges to label each generated sentence with the closest category.
We then calculate the normalized token-level edit distance between the judge-annotated label sequence and the given control sequence, where each label is considered as a single token, and finally deduct this value from 1.

For decision correctness, we evaluate it on a binary scale where 1 indicates complete correctness and 0 otherwise.
More specifically, we give 0 if the generation produces either contradictory decisions or a wrong decision, or if the generation does not show enough hints for rejection or acceptance. 

\end{document}